\documentclass[runningheads]{llncs}

\usepackage[T1]{fontenc}
\usepackage{graphicx}
\usepackage[export]{adjustbox}
\usepackage{bbding}

\usepackage{amsmath}
\usepackage{amssymb}

\usepackage{booktabs}
\usepackage{xcolor}     
\usepackage{rotating}
\usepackage{multirow}

\usepackage{siunitx}
\usepackage{pgfplots}
\pgfplotsset{compat=1.18}
\usetikzlibrary{positioning,calc,patterns,decorations.pathreplacing}
\usepgfplotslibrary{polar,groupplots}

\newcommand{\Ehat}{\hat{\mathcal{E}}}

\usepackage{hyperref}

\begin{document}

\title{DIAL-KG: Schema-Free Incremental Knowledge Graph Construction via Dynamic Schema Induction and Evolution-Intent Assessment} 
\titlerunning{DIAL-KG}

\author{Weidong Bao\thanks{Equal contribution} \and
Yilin Wang\protect\footnotemark[1] \and
Ruyu Gao \and
Fangling Leng \and
Yubin Bao(\Envelope) \and
Ge Yu} 
\authorrunning{W. Bao et al.}
\institute{Department of Computer Science and Engineering, Northeastern University, Shenyang, China\\
\email{\{baowd, gaory2\}@mails.neu.edu.cn, wangyilin0409@gmail.com, \\
\{lengfangling, baoyubin, yuge\}@cse.neu.edu.cn}} 
\maketitle

\begin{abstract}
Knowledge Graphs (KGs) are foundational to applications such as search, question answering, and recommendation. Conventional knowledge graph construction methods are predominantly static, relying on a single-step construction from a fixed corpus with a predefined schema. However, such methods are suboptimal for real-world scenarios where data arrives dynamically, as incorporating new information requires complete and computationally expensive graph reconstructions. Furthermore, predefined schemas hinder the flexibility of knowledge graph construction. To address these limitations, we introduce \textbf{DIAL-KG}, a closed-loop framework for incremental KG construction orchestrated by a \textbf{Meta-Knowledge Base (MKB)}. The framework operates in a three-stage cycle: (i) \textbf{Dual-Track Extraction}, which ensures knowledge completeness by defaulting to triple generation and switching to event extraction for complex knowledge; (ii) \textbf{Governance Adjudication}, which ensures the fidelity and currency of extracted facts to prevent hallucinations and knowledge staleness; and (iii) \textbf{Schema Evolution}, in which new schemas are induced from validated knowledge to guide subsequent construction cycles, and knowledge from the current round is incrementally applied to the existing KG. Extensive experiments demonstrate that our framework achieves state-of-the-art (SOTA) performance in the quality of both the constructed graph and the induced schemas.
\end{abstract}

\section{Introduction} \label{sec:introduction}
Knowledge Graphs (KGs) are foundational to applications such as search, question answering, and recommendation.\cite{distiawan2019neural,wu2024cotkr,chen2024new,lyu2023llm,huang2025alleviating,huang2025survey,huang2025improving}. Nonetheless, real-world knowledge is intrinsically complex, and conventional KGs construction(KGC) methods are heavily dependent on extensive manual effort\cite{hofer2023construction,ye2022generative,zhu2024llms,pan2024unifying}. 

\begin{figure}
  \centering
  \includegraphics[width=\linewidth]{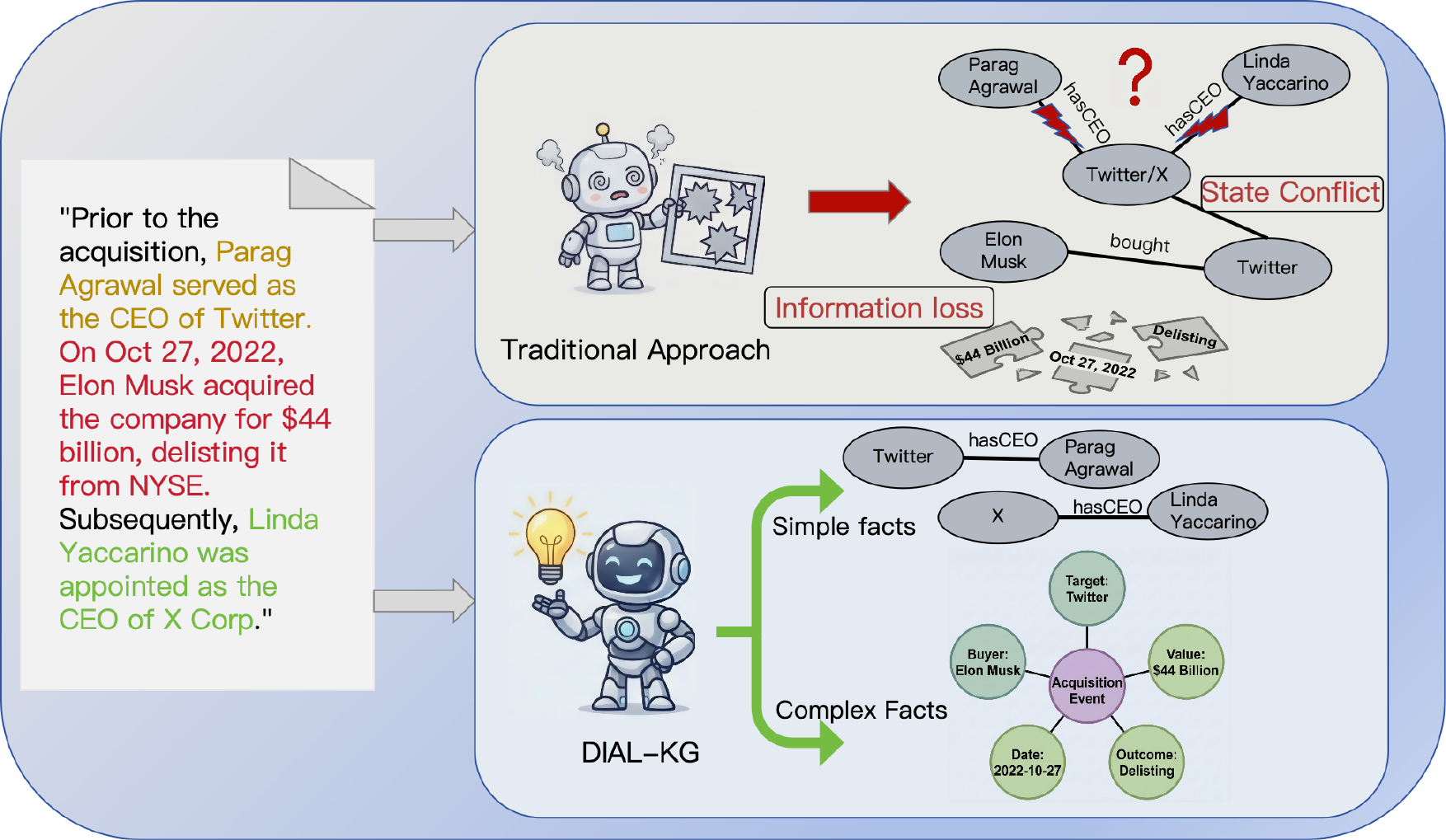}
  \caption{Comparison of extraction paradigms for dynamic and complex knowledge. Traditional methods (Top) rely on predefined static schemas, forcing the model (depicted as the overwhelmed agent) to compress dynamic histories and complex events into simple triples. This results in severe semantic distortion and temporal ambiguity. In contrast, DIAL-KG (Bottom) introduces a Dual-Track Extraction mechanism. By routing simple attribute changes (Static Track) and complex structural changes (Event Track) separately, DIAL-KG achieves schema-free evolution, ensuring both temporal clarity and comprehensive context preservation.}
  \label{fig:intro}
\end{figure}

Several approaches have been used to construct KGs. Rule-based systems apply predefined logical rules to extract and structure knowledge \cite{suchanek2007yago,lewis2019bart,hearst1992automatic,carlson2010toward,chiticariu2013rule}. They offer high precision and domain control but struggle with scalability, limited generalization, and fragility. Supervised models learn extraction patterns from annotated data\cite{suchanek2007yago,banko2008tradeoffs,raffel2020exploring}, yet face high annotation costs, limited adaptability, and dependence on training data. Large language models (LLMs) are increasingly used for automated extraction and graph construction
\cite{zhu2024llms,chen2023autokgefficientautomatedknowledge,xue2024autore,zhang2024extract,lairgi2024itext2kg}. However, many methods still rely on predefined schemas, as shown in Fig.~\ref{fig:intro},which restricts flexibility. Moreover, the absence of incremental lifecycle mechanisms impedes adding, modifying, and retiring knowledge while preserving traceability.

By drawing inspiration from human learning and cognitive correction mechanisms \cite{hassabis2017neuroscience}, we can more effectively perform incremental knowledge graph construction (IKGC) on dynamic data. Human learning is a process of continuous refinement: new knowledge integrates, adjusts, and extends prior understanding rather than overturning it. A physics student, for example, does not discard Newtonian mechanics after learning relativity but treats it as an approximation valid under specific conditions, with clearly defined boundaries within a broader theory. This relies on the human brain’s unique “versioned memory” mechanism, which preserves conclusions together with their provenance and revision history, including the evidence that prompted updates and the limits of earlier beliefs. Such a mechanism enables resilience and sustainable evolution when confronted with conflict. To replicate this evolutionary capacity, IKGC must handle knowledge in its inherent complexity. Because many statements exhibit \emph{n-ary} structure or \emph{time varying}~\cite{luo2024text2nkg,rosso2020beyond}, event representations are more appropriate; compressing them into triples discards the temporal and state signals needed to verify correctness\cite{galkin2020message,gottschalk2018eventkg,guo2022what,knez2023event,wadden2019entity}.

To address these challenges, we propose \textbf{DIAL-KG}, a closed-loop framework built around a Meta-Know-ledge Base (MKB) that serves as both governance hub and evolutionary memory. The MKB records, indexes, and updates meta-knowledge , including entity profiles, schema proposals that supplies batch-aware contextual constraints. With this context, the system conducts evidence verification, consistency checks, intent recognition, and auditable soft deprecation as transactional updates, enabling iterative, traceable, and reliable graph evolution without full reconstruction. Operationally, DIAL-KG executes a three stage loop for each batch $B_k$: (i) \textbf{Dual-Track Extraction}, which ensures knowledge completeness by defaulting to triple generation and switching to event extraction for complex knowledge. (ii) \textbf{Governance Adjudication}, which ensures the fidelity and currency of extracted facts to prevent hallucinations and knowledge staleness; and (iii) \textbf{Schema Evolution}, in which new schemas are induced from validated knowledge to guide subsequent construction cycles and knowledge from the current round is incrementally applied to the existing KG. 

The key contributions of this paper are:

\begin{itemize}
  \item \textbf{Closed-loop, incremental operation.} We recast KGC from static, open-loop pipelines into a governance-centric closed loop with transactional updates and soft deprecation, enabling auditable add, modify, and retire operations at batch granularity.

  \item \textbf{Evolution-aware representation with parsimony.} We establish a dual-track extraction regime that preserves time, status, and phase cues via events only when necessary, retaining sparsity while enabling lifecycle-aware decisions.

  \item \textbf{Self-evolving constraints via an MKB.} We operationalize a Meta-Knowledge Base that promotes relation and event schemas and consolidates entity profiles, turning them into retrieval constraints for later batches, thereby expanding coverage without rigid predefined schemas or heavy per-relation tuning.
\end{itemize}

We validate DIAL-KG on static benchmarks and a purpose-built streaming dataset, where it improves F1 by up to 4.7\% over strong schema-free LLM baselines, achieves over 98\% precision on evidence-backed soft deprecations in streaming, and delivers more compact schemas (up to 15\% fewer relation types) with a 1.6--2.8 point reduction in redundancy. These results confirm its superior performance in foundational extraction and, more importantly, its robustness in managing the incremental knowledge lifecycle with high fidelity.


\section{Related Work}

In recent years, the paradigm of Knowledge Graph  construction has undergone a profound shift from traditional rule-based and supervised learning to approaches driven by Large Language Models.

\subsubsection{Rule and supervised-learning--based KGC.}
The YAGO family of systems leverages hierarchical structures and prior knowledge for semantic extraction\cite{suchanek2007yago}. With advances in pre-trained language models such as T5 \cite{raffel2020exploring} and BART \cite{lewis2019bart}, subsequent work has framed Knowledge Graph Construction (KGC) as a sequence-to-sequence task, fine-tuning these models to generate relational triples end to end\cite{zhan2020span,banko2008tradeoffs}. However, this line of techniques generally relies on meticulous feature engineering or template design and substantial manual hyperparameter tuning, resulting in high cross-domain transfer costs, limited adaptability to long, complex documents with intertwined relations, and poor support for high-frequency updates and continual evolution.

\subsubsection{LLM-based KGC.}
The rise of LLMs has opened a disruptive path for automated KGC, which can be broadly organized into three lines. \textbf{Schema-/Ontology-guided.} These methods are anchored in external knowledge bases or domain ontologies (e.g., Wikipedia, DBpedia)\cite{volkel2006semantic,auer2007dbpedia} and leverage the generative and reasoning capabilities of LLMs to complete entity relations and types. For instance, SAC-KG retrieves domain corpora and DBpedia to provide context that guides triple generation\cite{chen2024sac}; CoT-Ontology combines domain ontologies with Chain-of-Thought (CoT) prompting for step-wise triple extraction\cite{mintz2009distant}; RAP proposes a schema-aware, retrieval-augmented approach that dynamically incorporates structured schema knowledge and semantically relevant instances as contextual prompts\cite{yao2023schema}. The main bottleneck is the strong dependence of construction quality and coverage on the completeness and accuracy of external ontologies: once these sources lag or contain bias, KG quality is constrained, and domain adaptation and maintenance become costly. \textbf{Fine-tuning-based.} This line of work, such as AutoRE\cite{xue2024autore}, aims to directly generate structured triples from unstructured documents, weakening the hard constraint of a ``predefined relation set.'' By updating a small subset of parameters, such methods can capture deep semantic relations in specific scenarios and deliver stable performance. However, as the relation space grows rapidly in practice, fine-tuning and maintenance complexity escalate, and generalization to novel relations and long-tail types remains limited.\textbf{Schema-free.} To eliminate reliance on predefined schemas, researchers explore more open pipelines that perform open information extraction first and induce/normalize schemas afterward. EDC decomposes KGC into three stages: open information extraction, schema definition, and schema normalization. However, its single-batch and static architecture exposes inherent limitations when applied to dynamic, continuous data streams~\cite{zhang2024extract}. iText2KG proposes a zero-shot, incremental construction method that resolves semantic ambiguity via user-defined blueprints; its core limitation, however, is the inability to proactively discover knowledge types beyond the blueprint’s scope and the lack of clear mechanisms for dynamic knowledge evolution (e.g., revising and deprecating outdated knowledge)\cite{lairgi2024itext2kg}. AutoKG\cite{chen2023autokgefficientautomatedknowledge} designs a multi-agent collaboration framework and integrates real-time web retrieval to build knowledge, fitting an agentic/RAG-centric paradigm that can complement schema-free extraction.

An effective dynamic KGC framework must combine automation, accuracy, adaptability, and incremental integration. DIAL-KG achieves this by unifying extraction, schema evolution, and lifecycle governance within a self-driven closed loop.

\section{Preliminaries}
\label{sec:preliminaries}

\subsection{Meta-Knowledge Base}
\label{sec:mkb}

The Meta-Knowledge Base (MKB) is the management core and evolving metadata repository of DIAL-KG. It comprises:
\textbf{Entity Profile}: a structured, normalized description of a real-world entity, consolidating verified canonical names, aliases, and types. Profiles act as semantic anchors for coreference resolution and entity alignment, ensuring consistency and contextual continuity over time.
\textbf{Schema Proposal}: Candidate schemas induced from accumulated, verified facts, corresponding to the dual-track extraction. They are categorized into: Relation Schemas: Define static fact structures, specifying the schema and its domain/range constraints. {Event Schemas}: Define dynamic event structures, specifying the event type, its trigger, and its argument roles with their constraints. Once validated, both types of schemas are integrated into the MKB to constrain subsequent extraction and logical validation (Sect.~\ref{sec:governance}), enabling the system to progressively form a self-evolving schema system without a fixed ontology.

\subsection{Task Definition: Incremental Knowledge Graph Construction}
\label{sec:task_definition}

We define Incremental Knowledge Graph Construction (IKGC) as a setting where text arrives continuously in streaming batches. At each timestep $k$, the system receives a new batch $B_k = \{d_1, \dots, d_n\}$. Given the existing graph state $G_{k-1}$ and the Meta-Knowledge Base $MKB_{k-1}$, the system autonomously produces the updated state:
\begin{equation}\label{eq:state-transition}
(G_{k-1}, MKB_{k-1}) \xrightarrow{\Phi(B_k)} (G_k, MKB_k),
\end{equation}
where $\Phi$ is the autonomous update function that extracts, validates, integrates, and generalizes knowledge from $B_k$.

\subsection{Knowledge Graph State Definition}
\label{sec:kg_state}

At a discrete timestep $k-1$, the knowledge graph is $G_{k-1} = (V_{k-1}, E_{k-1})$, where $V_{k-1}$ is the set of entity nodes and $E_{k-1}$ is the set of fact edges. Each edge $e \in E_{k-1}$ is a triple $\langle h, r, t \rangle$ indicating that relation $r$ connects head entity $h$ and tail entity $t$.

Each edge $e \in E_{k-1}$ has a traceable status mapping
\begin{equation}
s_{k-1}: E_{k-1} \to \{\textsc{Active}, \textsc{Deprecated}\},
\end{equation}
which indicates whether a fact is currently in effect. Outdated facts are not physically deleted. Instead, their status is set to \textsc{Deprecated} while retaining the associated evidence and timestamps. This design preserves historical evolution and enables \textbf{soft deprecation}.

\subsection{Entity and Event Normalization}
\label{sec:normalization}

Given the new batch $B_k$, let $M_k$ denote the set of raw entity mentions and $\mathcal{E}_k$ the set of raw event instances. Normalization for both entities and events proceeds in two steps—\emph{intra-batch} canonicalization followed by \emph{cross-batch} alignment.

\begin{align}
\underbrace{\tilde{M}_k}_{\text{intra-batch}} &= \pi_k^{\mathrm{intra}}(M_k), 
& 
\underbrace{\hat{M}_k}_{\text{cross-batch}} &= (\pi_k^{\mathrm{cross}}\circ \pi_k^{\mathrm{intra}})(M_k). \label{eq:entity_norm}
\end{align}

\begin{align}
\underbrace{\tilde{\mathcal{E}}_k}_{\text{intra-batch}} &= \psi_k^{\mathrm{intra}}(\mathcal{E}_k),
&
\underbrace{\hat{\mathcal{E}}_k}_{\text{cross-batch}} &= (\psi_k^{\mathrm{cross}}\circ \psi_k^{\mathrm{intra}})(\mathcal{E}_k). \label{eq:event_norm}
\end{align}

New entities in $\hat{M}_k$ are merged into historical nodes $V_{k-1}$ to yield $V_k$. The normalized events $\hat{\mathcal{E}}_k$ serve as the sole input to subsequent \emph{Event Relationalization} (Sect.~\ref{sec:schema_induction}).

\subsection{Knowledge Increment Definition}
\label{sec:increment_def}

At each update round, the system produces a knowledge increment
\begin{equation}\label{eq:delta}
\Delta G_k = (V_k^+, E_k^+, E_k^\downarrow),
\end{equation}
where $V_k^+$ are newly discovered entities, $E_k^+$ are verified new facts, and $E_k^\downarrow \subseteq E_{k-1}$ are facts identified for deprecation. Applying the increment yields $G_k = (V_k, E_k)$:
\begin{align}
V_k &= V_{k-1} \cup V_k^+, \label{eq:vk}\\
E_k &= (E_{k-1} \setminus E_k^\downarrow) \cup E_k^+. \label{eq:ek}
\end{align}
The status mapping $s_k$ is updated accordingly to maintain evolutionary consistency.

\begin{figure}[t]
  \centering
  \begin{adjustbox}{max size={\linewidth}{0.95\textheight}}
    \includegraphics{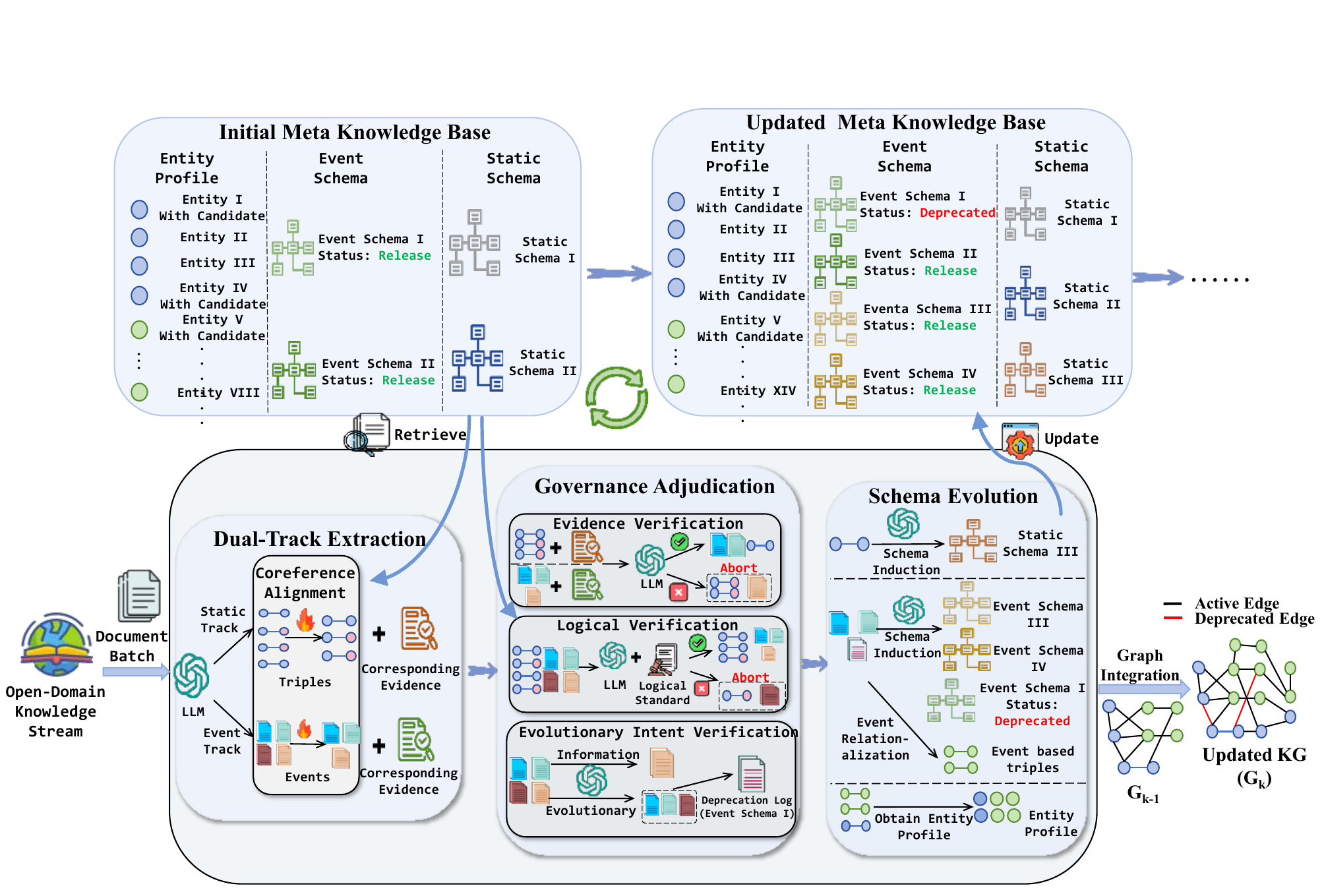}
  \end{adjustbox}
  \caption{Overall DIAL-KG framework.}
  \label{fig:dial}
\end{figure}

\section{Methodology}
\label{sec:methodology}

As shown in Fig.~\ref{fig:dial}, DIAL-KG constructs a dynamic knowledge graph through an autonomous, iterative process with three core stages:
(i) \textbf{Dual-Track Extraction}: in parallel, identify relation triples and event instances from streaming text;
(ii) \textbf{Governance Adjudication}: perform Evidence Verification, Logical Verification, and Evolutionary-Intent Verification with support from the MKB; and
(iii) \textbf{Schema Evolution}: generalize verified knowledge to induce new relation and event schemas.
Finally, \textbf{Transactional Integration} atomically applies $\Delta G_k$ to $G_{k-1}$ to produce $G_k$.

\subsection{Stage 1: Dual-Track Extraction}
\label{sec:extraction}

To balance parsimony with temporal fidelity, DIAL-KG dynamically routes input statements in batch $B_k$ to the appropriate representation track. Real-world statements differ in nature: stable, context-invariant assertions (e.g., ``Python is a programming language'') are encoded as relation triples $\langle h, r, t \rangle$, whereas complex statements involving timestamps, state transitions, or multi-argument semantics (e.g., ``In 2022, Microsoft announced Windows 10 EOL'') are represented as event structures $\epsilon = (\text{trigger}, \text{roles}, \text{time})$. This avoids over-structuring simple facts while preserving the nuances of evolving knowledge.

\textbf{MKB-Guided Generation.} The extraction mechanism adapts to the system state defined in Sect.~\ref{sec:task_definition}. In the cold-start phase ($k=0$), extraction relies on few-shot prompting. In the MKB-guided phase ($k > 0$), the system utilizes the \textit{Schema Proposals} $\mathcal{S}_{k-1}$ from the MKB. We vectorize the input to retrieve the top-$K$ relevant schemas from $\mathcal{S}_{k-1}$ (set to $K=30$ to balance schema recall with context length limits). These retrieved schemas are injected into the prompt as constraints. This retrieval-augmented generation ensures that new extractions remain consistent with the evolving schema ontology without exceeding the LLM's context window.

\begin{figure}[t]
  \centering
  \includegraphics[width=\linewidth]{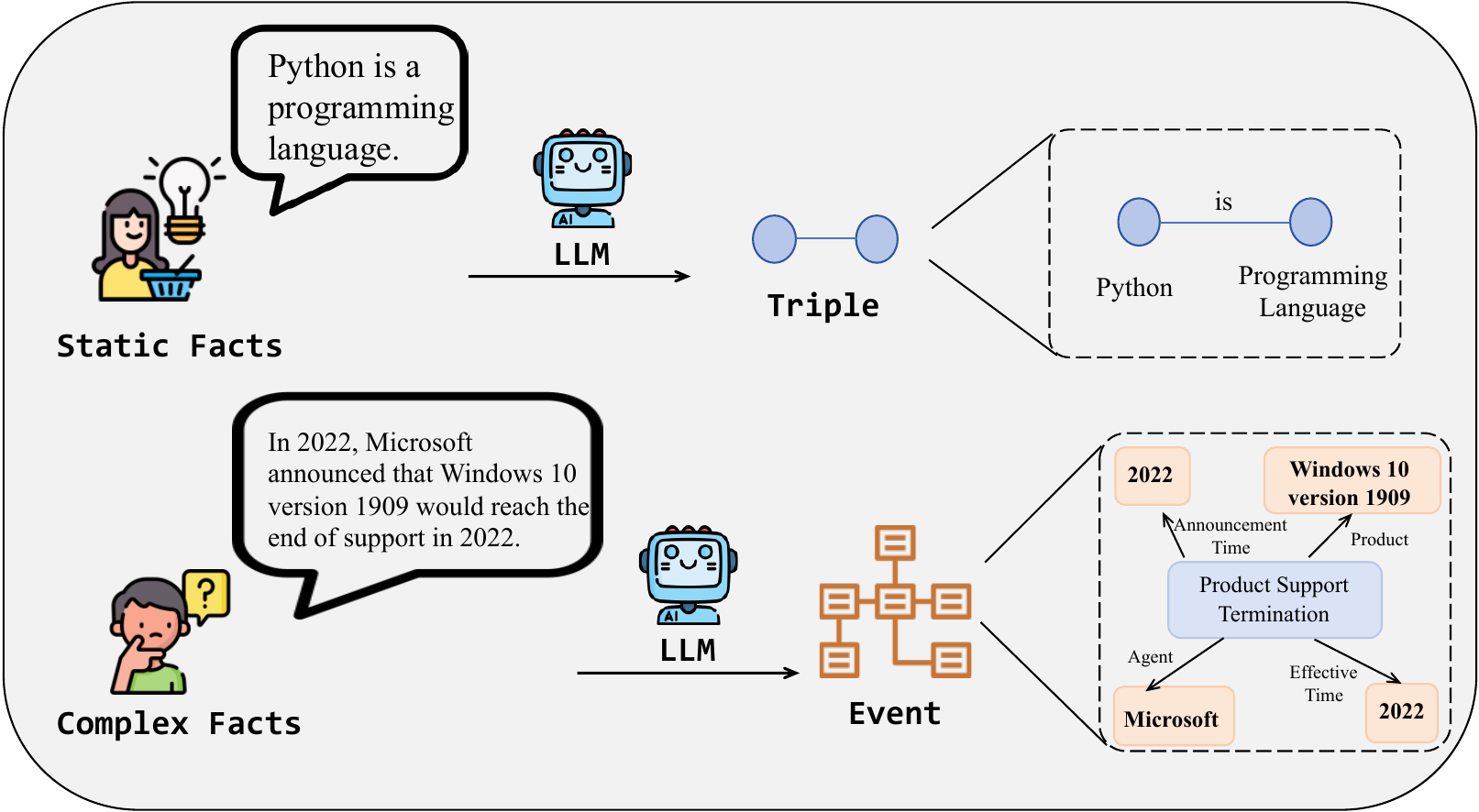}
  \caption{Dual-Track Extraction .}
  \label{fig:extract}
\end{figure}

\textbf{Coreference Alignment.}Coreference resolution operates at two levels to produce the normalized entities $\hat{M}_k$ and events $\hat{\mathcal{E}}_k$ defined in Sect.~\ref{sec:normalization}. In the cold-start phase ($k=0$), only intra-batch normalization is performed.

\textit{1)Intra-batch normalization.}For entities, we compute entity embeddings using an embedding model and cluster them based on embedding similarity, infer types with an LLM, and adjudicate pairs within same-type clusters using \{\textsc{Merge}, \textsc{Hierarchy}, \textsc{Separate}\}. This yields $\tilde{M}_k$ (the range of $\pi_k^{\mathrm{intra}}$). For events, characterized by trigger, argument set, and spatio-temporal constraints, we compute pairwise similarity, cluster candidates, and adjudicate clusters with the same decision set, yielding $\tilde{\mathcal{E}}_k$ (the range of $\psi_k^{\mathrm{intra}}$).

\textit{2)Cross-batch alignment.}Enabled when $k>0$, each entity in $\tilde{M}_k$ is matched to the MKB’s entity profiles to retrieve the top candidates. An LLM decides whether to reuse a historical ID or create a new one, completing $\pi_k^{\mathrm{cross}}$. For events, we query the MKB’s event index using trigger semantics, key arguments, and time window. If an event in $\tilde{\mathcal{E}}_k$ has a high match score and is consistent in time and arguments, it is aligned; otherwise, it is registered as new, completing $\psi_k^{\mathrm{cross}}$. The resulting $\hat{\mathcal{E}}_k$ is the sole input to Event Triplelization (Sect.~\ref{sec:schema_induction}) and supports Evolutionary-Intent Detection (Sect.~\ref{sec:governance}).

\subsection{Stage 2: Governance Adjudication}
\label{sec:governance}

This stage filters hallucinations and prevents knowledge staleness through three sub-processes: Evidence Verification, Logical Verification, and Evolutionary-Intent Verification.

\textbf{Evidence Verification.}Each extraction (triple or event) and its evidence segment are submitted to an LLM that judges strictly from the provided text, without external knowledge. A candidate is \textsc{Rejected} only if the evidence directly contradicts it; otherwise, it is conservatively \textsc{Accepted} to retain semantically correct candidates.

\textbf{Logical Verification.}Logical verification comprises
(i) general consistency checks that remove contradictions, e.g.,
\(\langle A, r, A\rangle\) with \(r=\mathrm{ancestor\_of}\),
or the mutually inverse pair
\(\langle A,\mathrm{part\_of},B\rangle\) and \(\langle B,\mathrm{part\_of},A\rangle\);
and (ii) schema-constraint checks (when \(k>0\) and MKB schemas exist) that
verify type signatures and event roles (e.g.,
\(\mathrm{ceo\_of}(\mathrm{Person},\mathrm{Organization})\)).
For example, if the \textsc{CorporateAcquisition} schema requires \texttt{acquirer}:\texttt{Organization},
then a candidate with \texttt{acquirer}~=~\texttt{Elon Musk (Person)} is rejected.

In the cold-start phase ($k=0$), only general consistency applies. When $k>0$, candidates matching known schemas must pass both criteria, whereas candidates for unseen schemas skip the schema-constraint check and proceed to dynamic induction (Sect.~\ref{sec:schema_induction}).

\textbf{Evolutionary-Intent Verification.}This process applies to normalized events $\epsilon \in \hat{\mathcal{E}}_k$ that pass the previous checks. The LLM assigns one of two intents: \textbf{Informational} (stating a fact, e.g., ``Google Inc. was founded on September~4, 1998, in Menlo Park'') or \textbf{Evolutionary} (indicating a state transition of historical knowledge, e.g., ``The company announced in 2022 it would discontinue support for product~X''). The system identifies evolutionary triggers (e.g., \emph{deprecated}, \emph{removed}, \emph{replaced}). Informational events contribute to $E_k^+$. For evolutionary events, the system retrieves the targeted outdated facts in $E_{k-1}$ and adds them to $E_k^\downarrow$, recording them in a deprecation log for final integration.

\subsection{Stage 3: Schema Evolution}
\label{sec:schema_induction}
Knowledge that passes the Governance Adjudication stage (including informational events and verified triples) enters this stage to induce new schemas and update the MKB.

\textbf{Relation Schema Induction}: Verified relation triples are clustered based on their relation $r$'s embedding (using cosine similarity). When a cluster's frequency exceeds a threshold $\theta$ and exhibits high semantic coherence, a relation schema candidate is generated. This proposal is evaluated by an LLM for semantic completeness and generalizability. Passed schemas are written to the MKB (with type signatures like domain/range, symmetric/anti-symmetric properties); failed ones are kept in a proposal pool for re-evaluation with more data.

\textbf{Event Schema Induction and Relationalization}: First, event schemas are induced. The system clusters instances from the normalized event set $\Ehat_k$ (represented by triggers, argument roles, time, etc.). Clusters meeting frequency and coherence thresholds generate an event schema candidate. After LLM evaluation, passed schemas are written to the MKB as formal event schemas (defining role sets and constraints). Subsequently, each normalized event instance in $\Ehat_k$ undergoes Relationalization for unified graph storage: a unique node $\epsilon$ is created for the event, and facts are generated, such as $\langle \epsilon, \text{rdf:type}, \text{EventType} \rangle$ and a series of argument facts $\langle \epsilon, \text{has\_role}, a_i \rangle$ (where $a_i$ is the subject, object, time, etc.).

\textbf{Entity Profile Update}: The system aggregates all verified new facts by entity ID, merges aliases, and normalizes key attributes. The updated $\text{EntityProfile}$ is written back to the MKB, and the vector index is synchronized to improve coreference resolution and extraction consistency in subsequent batches.

\subsection{Transactional Integration}
\label{sec:integration}
This final stage atomically applies all round-$k$ changes to the graph. The system aggregates $V_k^+$ (new entities from coreference resolution), $E_k^+$ (new facts verified in Governance Adjudication and facts from Event Triplelization), and $E_k^\downarrow$ (facts recorded in the Deprecation Log). The complete increment $\Delta G_k = (V_k^+, E_k^+, E_k^\downarrow)$ is applied to $G_{k-1}$ to obtain $G_k$ according to Sect.~\ref{sec:increment_def}, completing the incremental update.

\section{Experiments}
\label{sec:experiments}

\subsection{Experimental Setup}
\label{sec:setup}

\textbf{Implementation.} We employ \textbf{Qwen-Max} \cite{yang2025qwen3} for generation and reasoning tasks, and leverage \textbf{DeepSeek-V3} \cite{liu2024deepseek} (temp=0.1) as an independent judge \cite{chen2024sac,zheng2023judging}. Semantic similarity is computed via \textbf{BGE-M3}\footnote{\url{https://huggingface.co/BAAI/bge-m3}}.

\textbf{Datasets.} We use two static benchmarks, \textbf{WebNLG}\cite{ferreira20202020} and \textbf{Wiki-NRE}\cite{distiawan2019neural}, adapted for streaming via deterministic slicing. Crucially, we construct \textbf{SoftRel-$\Delta$} (1,515 entries) from Kubernetes release logs across three windows ($\Delta_1$: Baseline, $\Delta_2$: Evolution signals, $\Delta_3$: Consolidation). \emph{While distinct from real-time news, this windowed setup rigorously isolates and tests the system's ability to handle deprecation and evolution.}

\textbf{Baselines.} We compare against two state-of-the-art schema-free LLM extractors: \textbf{EDC} \cite{zhang2024extract} and \textbf{AutoKG} \cite{chen2023autokgefficientautomatedknowledge}. Note that traditional Temporal KGC methods are excluded as they rely on predefined schemas, contradicting our schema-free setting.

\textbf{Research Questions.} Our evaluation addresses three core questions: \textbf{RQ1 (Static Quality)} investigates how \textsc{DIAL-KG} compares with baselines on foundational extraction tasks; \textbf{RQ2 (Incremental Reliability)} assesses whether the system can reliably add new facts and deprecate obsolete ones in streaming settings; and \textbf{RQ3 (Schema Quality)} evaluates if the induced schema maintains compactness and minimizes redundancy.

\textbf{Metrics.} For static tasks, we report standard Precision, Recall, and F1. For streaming (RQ2), we introduce two incremental metrics for window $t$:
(1) \textbf{$\Delta$-Precision:} The accuracy of newly added facts $A_t$, defined as $\Delta\text{-P}_t = |TP_t| / |A_t|$, where $TP_t$ denotes additions judged as \texttt{fully\_supported} by the LLM.
(2) \textbf{Deprecation-Handling Precision (D-HP):} The reliability of soft deprecations $\mathcal{D}_t$, calculated as $\text{D-HP}_t = |JD_t| / |\mathcal{D}_t|$, where $JD_t$ represents deprecations supported by explicit textual evidence.

\begin{table}[htbp]
\centering
\caption{Static extraction performance on \textit{WebNLG}, \textit{Wiki-NRE}, and \textit{SoftRel-$\Delta$}.
Baselines vs.\ \textsc{DIAL-KG} in \emph{Batch-Static} and \emph{Stream-End-Static} (static scoring at the end).}
\label{tab:static_joint}
\begin{tabular}{ll
                S[table-format=1.3]
                S[table-format=1.3]
                S[table-format=1.3]}
\toprule
\textbf{Dataset} & \textbf{Model (Mode)} & {\textbf{Precision}} & {\textbf{Recall}} & {\textbf{F1-Score}} \\
\midrule
\multirow{4}{*}{\textit{WebNLG}}
& EDC (baseline)                 & 0.835 & 0.862 & 0.848 \\
& AutoKG (baseline)              & 0.781 & 0.801 & 0.791 \\
& \textbf{DIAL-KG (Batch)}       & \bfseries 0.848 & \bfseries 0.883 & \bfseries 0.865 \\
& \textbf{DIAL-KG (Stream-End)}  & 0.842 & 0.872 & 0.857 \\
\midrule
\multirow{4}{*}{\textit{Wiki-NRE}}
& EDC (baseline)                 & 0.784 & 0.833 & 0.808 \\
& AutoKG (baseline)              & 0.792 & 0.840 & 0.815 \\
& \textbf{DIAL-KG (Batch)}       & \bfseries 0.822 & \bfseries 0.887 & \bfseries 0.853 \\
& \textbf{DIAL-KG (Stream-End)}  & 0.814 & 0.876 & 0.844 \\
\midrule
\multirow{4}{*}{\textit{SoftRel-$\Delta$}}
& EDC (baseline)                 & 0.901 & 0.892 & 0.897 \\
& AutoKG (baseline)              & 0.894 & 0.887 & 0.891 \\
& \textbf{DIAL-KG (Batch)}       & \bfseries 0.933 & \bfseries 0.909 & \bfseries 0.922 \\
& \textbf{DIAL-KG (Stream-End)}  & 0.931 & 0.910 & 0.920 \\
\bottomrule
\end{tabular}
\end{table}

\subsection{Results}
\label{sec:main_results}

\noindent\textbf{Static Performance.} Table~\ref{tab:static_joint} shows that \textsc{DIAL-KG} in \emph{Batch-Static} consistently surpasses static baselines, indicating a strong foundational extractor. In \emph{Stream-End-Static}, performance on \textit{WebNLG} and \textit{Wiki-NRE} remains within 1–2 F1 points of the batch setting, reflecting mild cold-start and sequencing effects. On the evolutionary \textit{SoftRel-$\Delta$} dataset, the batch–stream gap is negligible: thematically linked windows allow memory and the MKB to recover or refine facts later, validating robustness under realistic information evolution.

\begin{table}[htbp]
\centering
\caption{Incremental decision quality on \textit{WebNLG}, \textit{Wiki-NRE}, and \textit{SoftRel-$\Delta$}.
We report $\Delta$-Precision for additions and D-HP for soft deprecations (N/A where no deprecations occur).}
\label{tab:incremental_main}
\begin{tabular}{l l
                S[table-format=1.3]
                l}
\toprule
\textbf{Dataset} & \textbf{$\Delta$} & {\textbf{$\Delta$-Precision}} & \textbf{D-HP} \\
\midrule
\multirow{2}{*}{\textit{WebNLG}}
& $\Delta_2$ & \bfseries 0.975 & N/A \\
& $\Delta_3$ & \bfseries 0.976 & N/A \\
\midrule
\multirow{2}{*}{\textit{Wiki-NRE}}
& $\Delta_2$ & \bfseries 0.972 & N/A \\
& $\Delta_3$ & \bfseries 0.974 & N/A \\
\midrule
\multirow{2}{*}{\textit{SoftRel-$\Delta$}}
& $\Delta_2$ & \bfseries 0.978 & \bfseries 0.986 \\
& $\Delta_3$ & \bfseries 0.973 & \bfseries 0.983 \\
\bottomrule
\end{tabular}
\end{table}

\noindent\textbf{Incremental Reliability.} Table~\ref{tab:incremental_main} reports per-window $\Delta$-Precision (additions) and D-HP (soft deprecations). \textit{WebNLG} and \textit{Wiki-NRE} contain no explicit deprecations (D-HP = N/A). On \textit{SoftRel-$\Delta$}, D-HP $>\!0.98$ indicates that deprecations are executed only when textually justified. Across open-domain (\textit{WebNLG}), Wikipedia-style (\textit{Wiki-NRE}), and domain-specific (\textit{SoftRel-$\Delta$}) corpora, $\Delta$-Precision $\ge\!0.97$ demonstrates robust online decisions; later windows further benefit from schemas/entities accumulated after the $\Delta_1$ cold start.

\begin{figure} 
  \centering
  \includegraphics[width=0.8\linewidth]{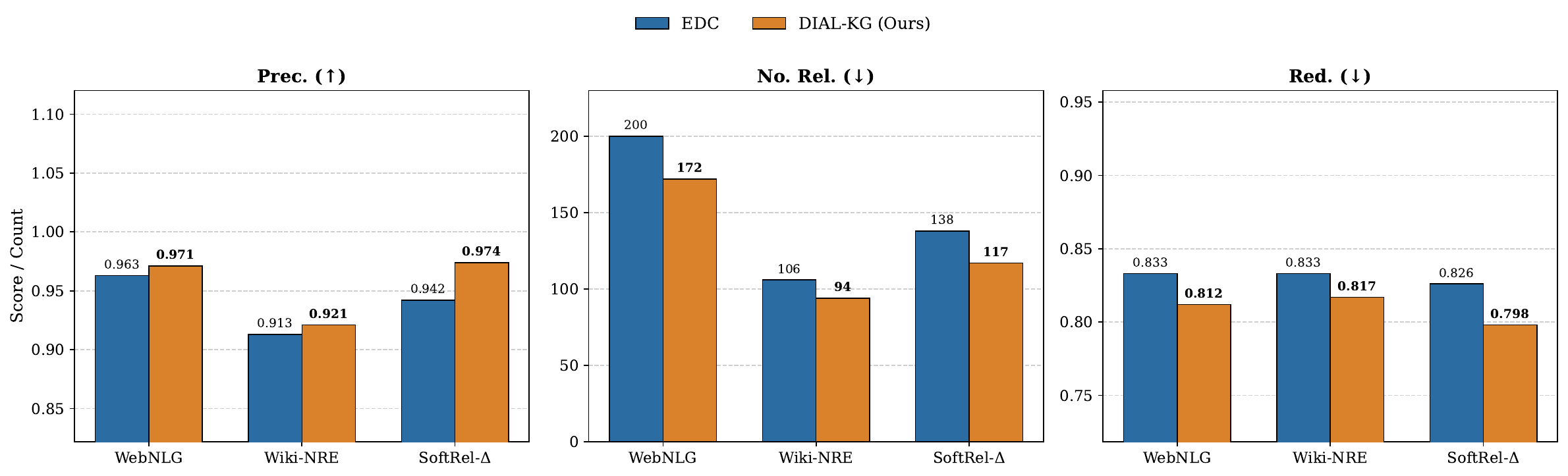}
  \caption{Schema quality comparison. \textsc{DIAL-KG} achieves higher precision with lower redundancy.}
  \label{fig:schema-quality2}
  \vspace{-5mm} 
\end{figure}

\noindent\textbf{Schema Quality.} \textsc{DIAL-KG} consistently outperforms EDC across all datasets (Fig.~\ref{fig:schema-quality2}). It achieves higher precision (an improvement of 0.8–3.2 points) while producing more compact schemas, with up to 15\% fewer relation types and a 1.6–2.8 point reduction in redundancy. EDC often generates near-duplicate relations such as \texttt{acquired\_by} and \texttt{acquisition\_of}, whereas \textsc{DIAL-KG} consolidates them into unified predicates through cross-batch canonicalization. These results indicate that DIAL-KG produces accurate, compact, and non-redundant schemas through self-canonicalized schema evolution.

\subsection{Ablation Study \& Case Study}
\label{sec:ablation_case}

\textbf{Ablation Analysis.} Table~\ref{tab:ablation} confirms the necessity of our core modules. \textsc{DIAL-KG}'s core contributions target streaming; thus, cross-window signals necessary for (i) Evolution-Intent Assessment (EIA), (ii) event representations, and (iii) Coreference Alignment are largely absent in \emph{Batch-Static}. We therefore ablate on the incremental metrics ($\Delta$-Precision, D-HP).
Removing EIA eliminates dependable deprecations (D-HP = N/A). Ablating event representations harms decisions on multi-argument facts. This confirms that event representation and intent assessment are indispensable for reliable streaming governance. Disabling coreference alignment degrades D-HP (to $\approx\!0.33$) due to entity fragmentation, which hinders precise targeting of historical facts for deprecation.

\begin{table}[ht] 
\vspace{-5mm} 

\centering
\caption{Ablation on \textit{SoftRel-$\Delta$}. ``N/A'' indicates missing functional capabilities.}
\label{tab:ablation}
\small 
\setlength{\tabcolsep}{5pt}
\vspace{-2mm} 
\renewcommand{\arraystretch}{0.95}
\begin{tabular}{l c c}
\toprule
\textbf{Ablation Variant} & \textbf{$\Delta$-Precision} & \textbf{D-HP} \\
\midrule
\textbf{Full Model} & \textbf{0.976} & \textbf{0.985} \\
w/o Intent Assessment & 0.848 & N/A \\
w/o Event Representation & 0.850 & N/A \\
w/o Coreference Alignment & 0.860 & 0.322 \\
\bottomrule
\end{tabular}
\vspace{-4mm} 
\end{table}

\textbf{Case Study: Lifecycle Governance.} To illustrate the system's handling of evolution, we present a representative case from \textit{SoftRel-$\Delta$} below.

\begin{center}
\setlength{\fboxsep}{6pt}
\setlength{\fboxrule}{0.6pt}
\fbox{\parbox{0.95\linewidth}{
    \small
    \textbf{Input Stream ($\Delta_2$):} ``\textit{The PodSecurityPolicy API is deprecated in v1.21 and will be removed in v1.25.}''
    \par\vspace{3pt} 
    \hrule height 0.4pt
    \par\vspace{3pt}
    \textbf{1. Dual-Track Extraction:} The system extracts an Event: \texttt{\{Trigger: deprecated, Target: PodSecurityPolicy\}}. \\
    \textbf{2. Intent Verification:} The LLM identifies the intent as \texttt{Evolutionary} rather than purely Informational. \\
    \textbf{3. MKB Governance:} The system queries the MKB for existing relations involving \texttt{PodSecurityPolicy}. It adds the new fact \texttt{(PodSecurityPolicy, status, deprecated)} and executes a \textbf{Soft Deprecation} on the outdated fact \texttt{(PodSecurityPolicy, status, active)}.
}}
\end{center}

\subsection{Judge Reliability and Validity Controls}
\label{sec:judge_reliability}
We fix the judge prompt and set temperature to $0.1$. For $\Delta$-Precision, only \texttt{fully\_supported} counts as correct; \texttt{partially\_supported} and \texttt{not\_supported} count as errors. For D-HP, a deprecation is correct only if the judge returns \texttt{deletion\_justified=true} with explicit evidence. Manual spot checks show substantial agreement; disagreements concentrate on borderline paraphrases rather than contradictions.

\section{Conclusion}

We propose \textsc{DIAL-KG}, a schema-agnostic, closed-loop architecture for incremental knowledge graph (KG) construction, anchored by an evolving Meta-Knowledge Base (MKB). By synergizing dual-track extraction, governance adjudication, and dynamic schema evolution, the framework facilitates autonomous schema induction and lifecycle management without necessitating a predefined ontology. Our empirical evaluations demonstrate that \textsc{DIAL-KG} achieves high-fidelity extraction and robust incremental updates through an auditable transactional mechanism. While the current reliance on Large Language Models (LLMs) for governance introduces latency constraints in high-velocity data environments, we contend that this computational overhead represents a strategic trade-off. Unlike conventional paradigms that require exhaustive graph reconstructions to incorporate new information, \textsc{DIAL-KG} substantially reduces the long-term amortized cost of KG maintenance. Future research will explore alleviating latency bottlenecks via the distillation of MKB-resident knowledge into specialized small language models (SLMs) and extending the framework to encompass multimodal signals for a more holistic knowledge evolution.

\subsubsection*{Acknowledgments} 
This research was funded by the National Natural Science Foundation of China (Nos. 62272093, 62137001).



\bibliographystyle{splncs04}
\bibliography{refs}
\end{document}